\documentclass[conference]{IEEEtran}
\IEEEoverridecommandlockouts
\usepackage{cite}
\usepackage{amsmath,amssymb,amsfonts}
\usepackage{algorithmic}
\usepackage{graphicx}
\usepackage{textcomp}
\usepackage{xcolor}

\usepackage{stfloats}
\usepackage{caption}
\usepackage{subcaption}
\usepackage{booktabs}
\usepackage{soul}
\usepackage[flushleft]{threeparttable}

\DeclareMathOperator{\Tr}{Tr}

\def\BibTeX{{\rm B\kern-.05em{\sc i\kern-.025em b}\kern-.08em
    T\kern-.1667em\lower.7ex\hbox{E}\kern-.125emX}}
    
\newcommand{\linebreakand}
{
  \end{@IEEEauthorhalign}
  \hfill\mbox{}\par
  \mbox{}\hfill\begin{@IEEEauthorhalign}
}

\DeclareMathOperator*{\argmin}{arg\,min}
    
\begin{document}
\bstctlcite{BibControl}

\title{
{Experimental Evaluation of Pose Initialization Methods for Relative Navigation Between Non-Cooperative Satellites}
\thanks{Project BIRD181070 funded by the program BIRD 2018 sponsored by the University of Padova.}
}

\author{
\IEEEauthorblockN{Sebastiano Chiodini}
\IEEEauthorblockA{\textit{CISAS ``Giuseppe Colombo''} \\
\textit{University of Padova}\\
Padova, Italy \\
sebastiano.chiodini@unipd.it}
\and
\IEEEauthorblockN{Marco Pertile}
\IEEEauthorblockA{\textit{Dept. of Industrial Engineering} \\
\textit{University of Padova}\\
Padova, Italy \\
marco.pertile@unipd.it}
\and
\IEEEauthorblockN{Pierdomenico Fracchiolla}
\IEEEauthorblockA{\textit{School of Engineering} \\
\textit{University of Padova}\\
Padova, Italy \\
pierdomenico.fracchiolla@studenti.unipd.it}
\and
\IEEEauthorblockN{Andrea Valmorbida}
\IEEEauthorblockA{\textit{Dept. of Industrial Engineering} \\
\textit{University of Padova}\\
Padova, Italy\\
andrea.valmorbida@unipd.it}
\and
\IEEEauthorblockN{Enrico Lorenzini}
\IEEEauthorblockA{\textit{Dept. of Industrial Engineering} \\
\textit{University of Padova}\\
Padova, Italy\\
enrico.lorenzini@unipd.it}
\and
\IEEEauthorblockN{Stefano Debei}
\IEEEauthorblockA{\textit{Dept. of Industrial Engineering} \\
\textit{University of Padova}\\
Padova, Italy\\
stefano.debei@unipd.it}
}

\maketitle

\begin{abstract}
In this work, we have analyzed the problem of relative pose initialization between two satellites: a chaser and a non-cooperating target. The analysis has been targeted to two close-range methods based on a monocular camera system: the Sharma-Ventura-D’Amico (SVD) method
and the silhouette matching method. Both methods are based on a priori knowledge of the target geometry, but neither fiducial markers nor a priori range measurements or state information are needed. The tests were carried out using a 2U CubeSat mock-up as target attached to a motorized rotary stage to simulate its relative motion with respect to the chaser camera. A motion capture system was used as a reference instrument that provides the fiducial relative motion between the two mock-ups and allows to evaluate the performances of the initialization algorithms analyzed.
\end{abstract}


\section{Introduction}

Accurate relative pose estimation between two spacecraft (SCs) are required for several scenarios, such as autonomous rendezvous and docking for on-orbit servicing, formation flight of two or more SCs, and active debris removal.
A recent and comprehensive review of methods for relative pose measurement is presented in \cite{opromolla2017review}. Tracking of fiducial and infrared markers integrated on the surface of the target SC has been demonstrated to be a reliable method for pose estimation as they are easily detectable with thermal imaging cameras even under extreme lighting conditions. This technique has proven to work in the case of ESA Automated Transfer Vehicle (ATV) in the range from 5 to 50 m \cite{medina2017towards}. A metrological characterization of a vision-based system for relative pose measurements based on fiducial markers is presented in \cite{pertile2018metrological}.
In \cite{tweddle2015relative}, Tweddle et al. present a relative navigation system based on fiducial markers and targeted to small inspection spacecrafts; this system has been tested on the Synchronized Position Hold Engage and Reorient
Experimental Satellites (SPHERES) testbed of the MIT. Astrobee \cite{bualat2018astrobee}, the new free-flying robot that operates on board the ISS, uses fiducial markers for autonomous docking, obtaining an accuracy of the order of one centimeter. The authors of \cite{sansone2018relative} presents a relative navigation technique based on LED fiducial markers. 

\begin{figure}[h!]
    \centering
   
        \subfloat[Sharma-Ventura-D’Amico (SVD) method]{\includegraphics[width=0.4\textwidth]{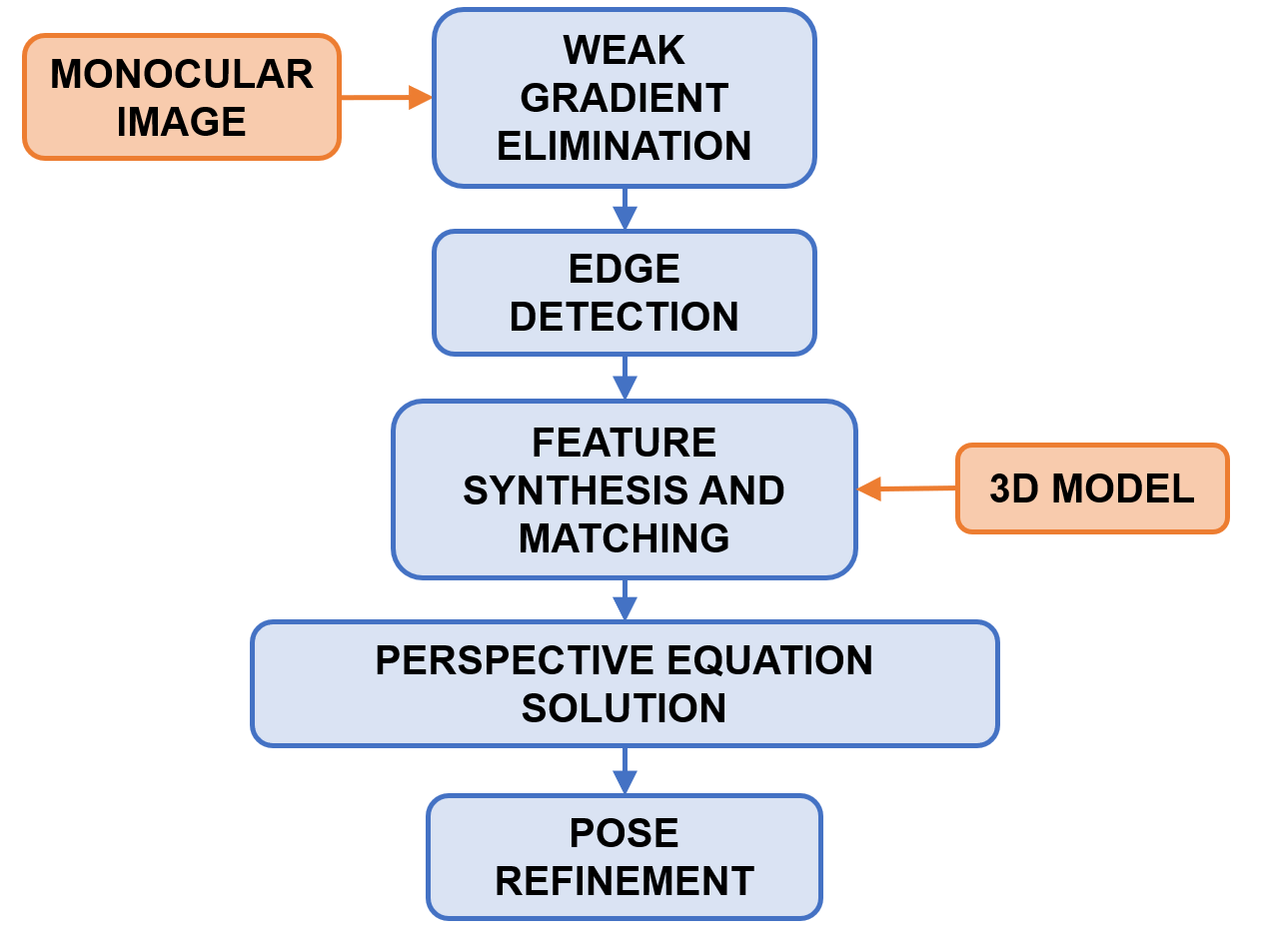}\label{fig:schemaSVD}}
        \vfill
        \subfloat[Silhouette
matching method]{\includegraphics[width=0.4\textwidth]{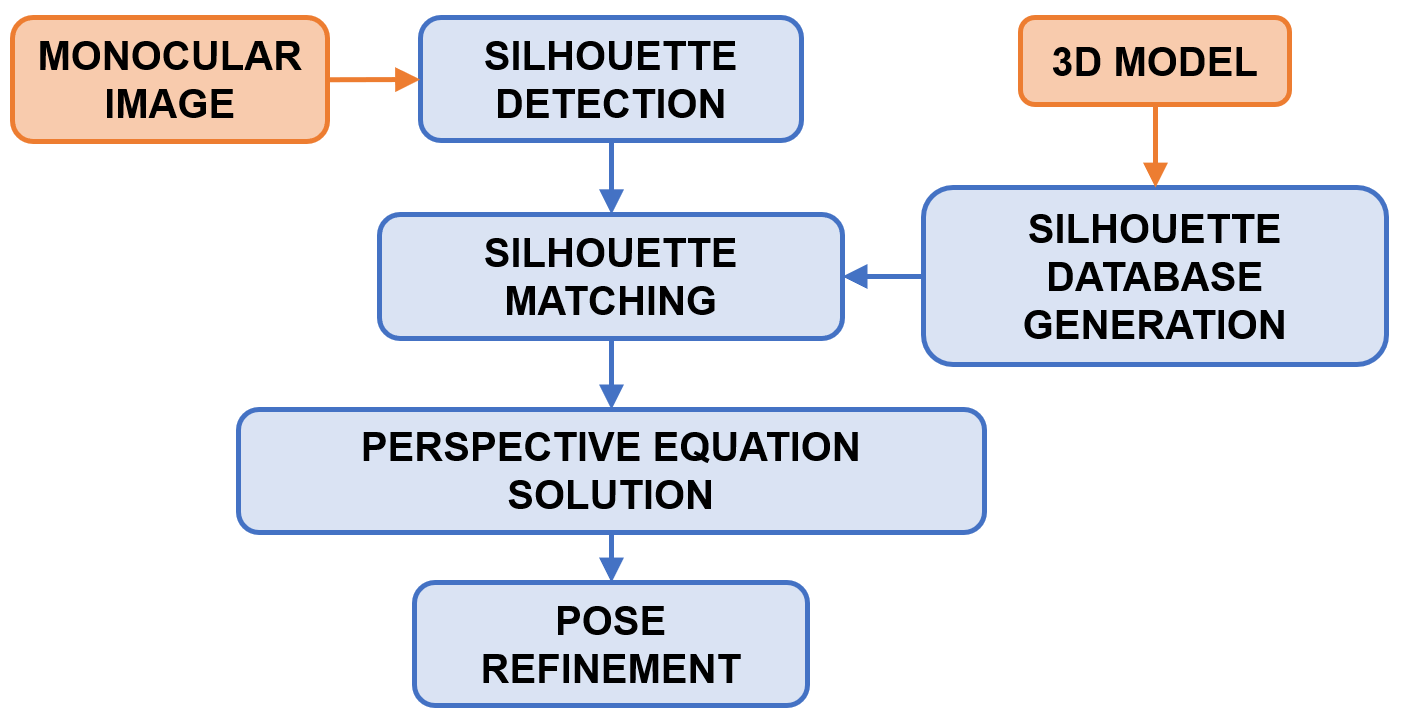}\label{fig:schemaSilhouett}}
   
    \caption{ Pose initialization methods processing flow. Input elements, highlighted by red rectangles, are processed by algorithm function, highlighted by blue rectangles.
}
\label{fig:methodsSchemes}
    
\end{figure}

\begin{figure*}[ht!]
    \centering
   
        \subfloat[]{\includegraphics[width=0.3\textwidth]{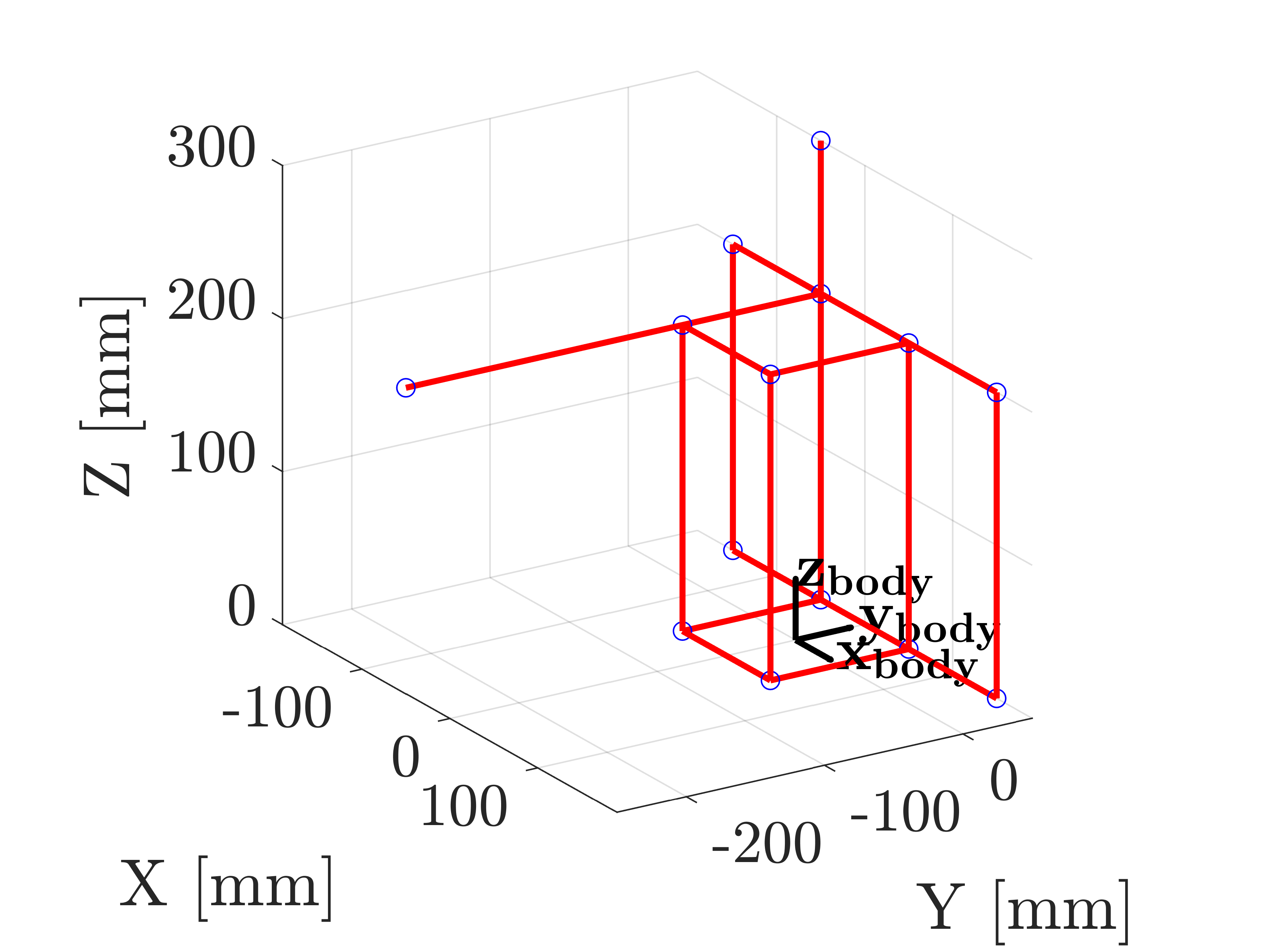}\label{fig:satelliteWireFrame}}
        \hfill
        \subfloat[]{\includegraphics[width=0.7\textwidth]{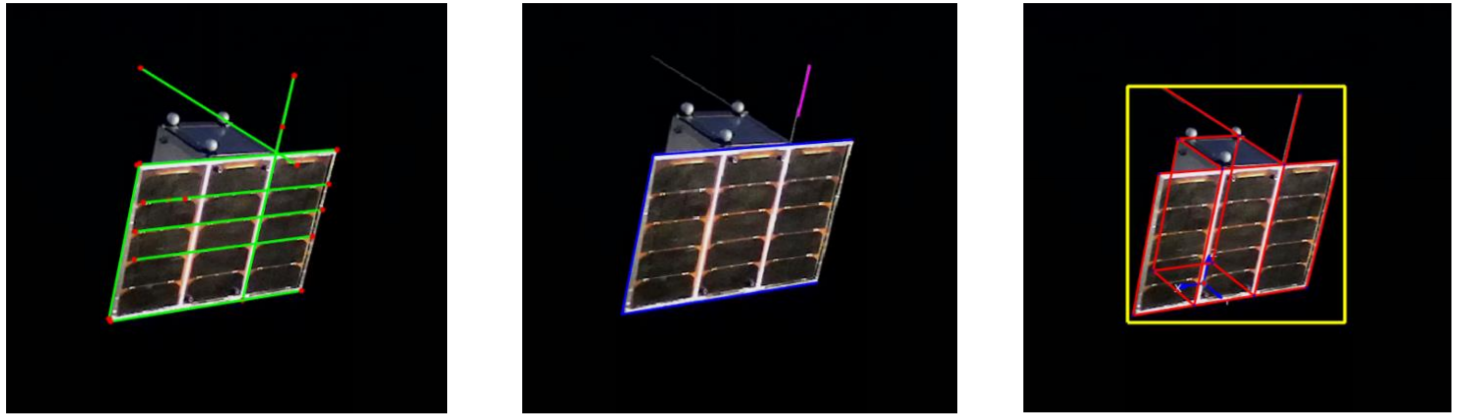}\label{fig:recognizedSatellite}}
   
    \caption{(a) Wireframe model of the CubeSat mock-up. (b) Results of SVD method on the CubeSat mock-up used during testing. From left to right, the
detected lines, the couple of polygonal triad and antennas that produce the optimal solution and the
model projected on the image after correct pose estimation.
}
    
\end{figure*}

The available space  on the SC surface may not always allow allocating of such markers, or in case of an existing SC removal scenario such markers have never been arranged, hence the need to use marker-less navigation techniques.
To provide range information, vision systems can be combined with a stereo system which, however, (1) increases the computational resources required for stereo-image processing, and (2) is subject to high depth uncertainties due to the small stereo baseline that can be accommodated in a satellite and the great distances involved. Another option is to use time-of-flight measurement systems, which are typically bulkier and more expensive. Pose initialization with a monocular system of known geometry targets seems to be a promising solution as it is a lightweight, inexpensive, and computationally undemanding system. Once the pose is initialized, it is possible to track the target SC with pose tracking methods such as SLAM or Kalman Filter \cite{dor2018orb}. Moreover, pose initialization on a geometric target with known geometry has an additional advantage as all target SC appendices that have to be manipulated by the chaser SC are expressed in a known reference frame.

Single image satellite pose estimation algorithms based on Convolutional Neural Network (CNN) architectures are giving very promising results, the authors of \cite{park2019towards} show a solution where the CNN network is trained so that the extracted 2D keypoints correspond to the points of a 3D model in pre-defined order. To improve the accuracy of the pose estimation the authors of \cite{pasqualetto2020cnn} associate a covariance matrix to the keypoints location returned by the CNN for each detected feature. However, to train the neural networks, a large amount of images are needed, which are often generated by computer graphics rendering due to the lack of real images, this led to a gap between real and simulated performances.

Ground testbed for optical navigation can be very useful both for algorithms validation and neural networks training. Recent development examples are the Robotic Testbed for Rendezvous and Optical Navigation (TRON) \cite{park2021} at Stanford University and the ESA’s GNC Rendezvous, Approach and Landing Simulator (GRALS) facility \cite{PASQUALETTOCASSINIS2022123}. To obtain the groud truth for the relative motion between satellite mock-ups, methods based on motion capture are widely used: these methods perform the tracking of a series of retroreflective markers placed on the satellite mock-ups. Dedicated calibration procedures are required to align the tracked system to the camera's reference system, as described in \cite{chiodini2018camera}, \cite{chiodini2019experimental} and \cite{valmorbida2020calibration}.

In this paper, we address the problem of pose initialization for non-cooperative satellite proximity operations of a target with known geometry. Two close-range methods are described and analysed: the Sharma-Ventura-D'Amico (SVD) method \cite{sharma2018robust}, which is based on finding corresponding geometric features (such as parallel segments, triads and antennas) between a known three-dimensional wireframe model and the target image, and the silhouette matching method \cite{petit20153d}. When compared to methods based on natural networks, purely geometric methods have the advantage of not requiring any training phase: the knowledge of the geometric model of the target satellite is sufficient. Furthermore, such methods can be used to perform unsupervised training of CNN-based methods.

The two methods are evaluated in terms of converged estimates and ground truth error by processing a dataset of images built from a 2U CubeSat mock-up. The CubeSat was rotated on a rotary stage to simulate the relative motion between the chaser camera and the target. Moreover, the camera-target relative motion was recorded by means of a motion capture system.

In Section \ref{sec:methods} the analysed methods, the SVD and the silhoutte method, are presented. In Section \ref{sec:results} the experimental set-up and the results of the experimental are presented. In Section \ref{sec:conclusions} the final remarks are reported.

\begin{figure*}[ht!]
    \centering
   
        \subfloat[]{\includegraphics[height=0.28\textwidth]{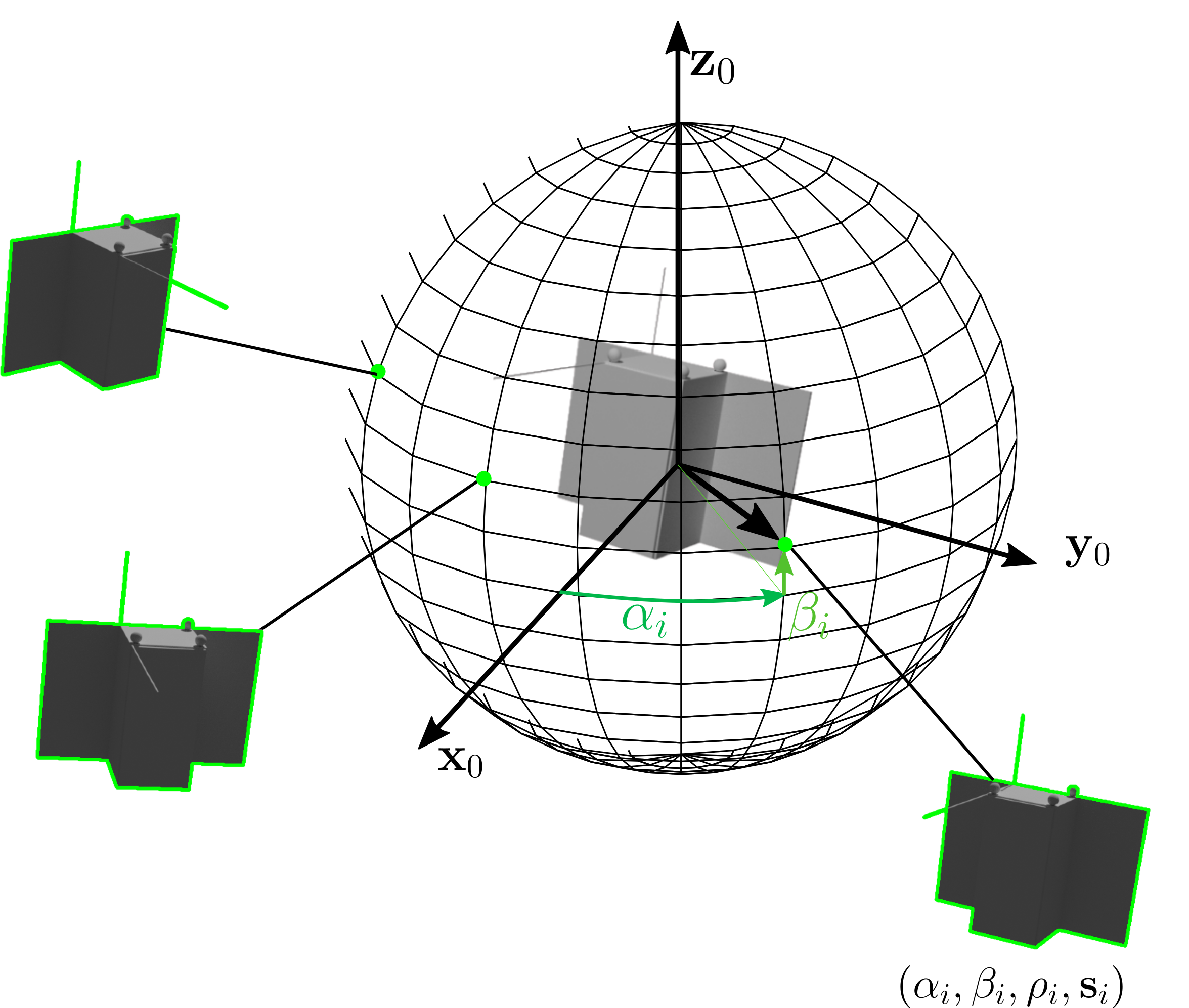}\label{fig:shiluhetteGeneration}}
        \hfill
        \subfloat[]{\includegraphics[height=0.28\textwidth]{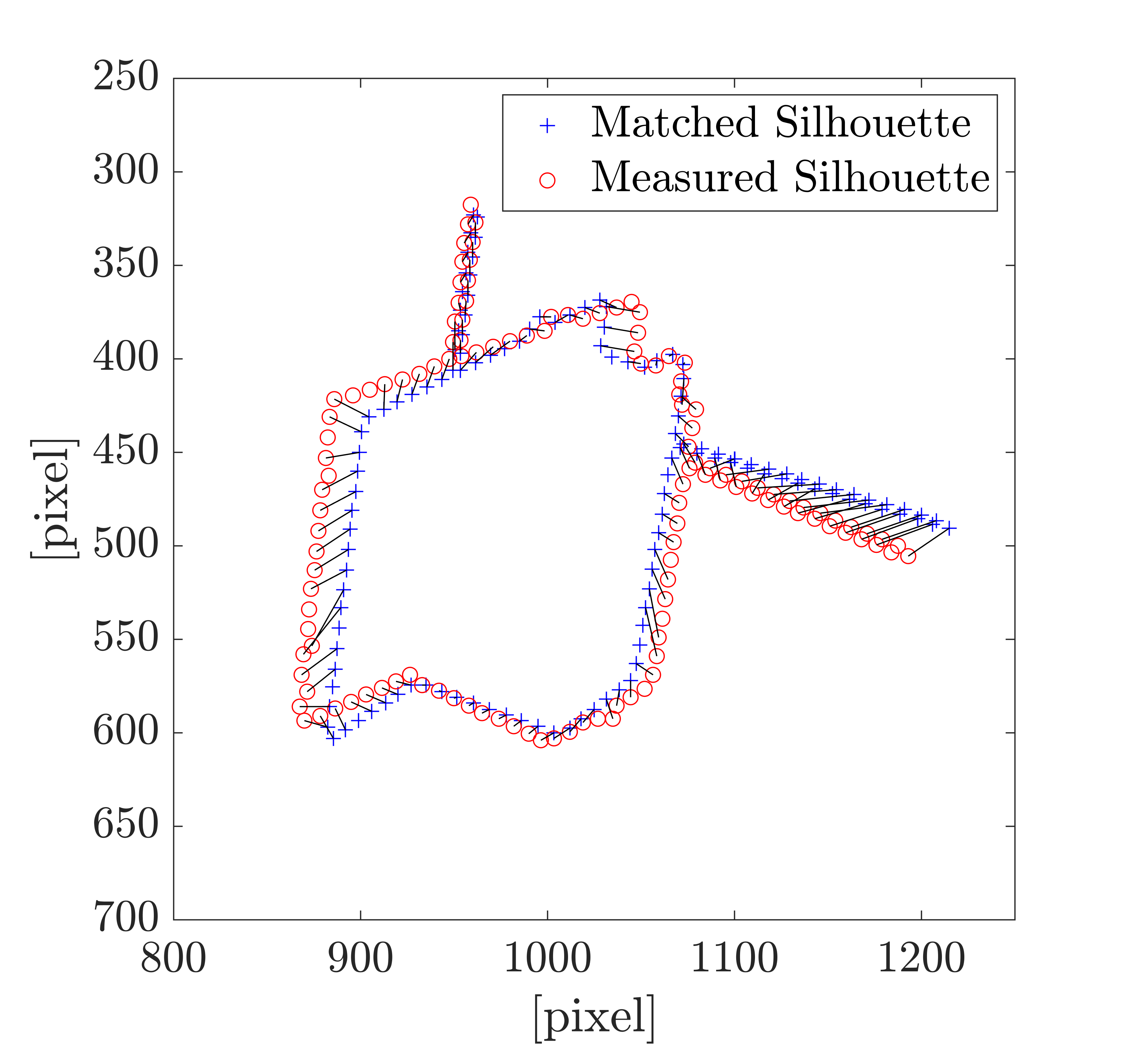}\label{fig:countourCorrespondences}}
        \hfill
        \subfloat[]{\includegraphics[height=0.28\textwidth]{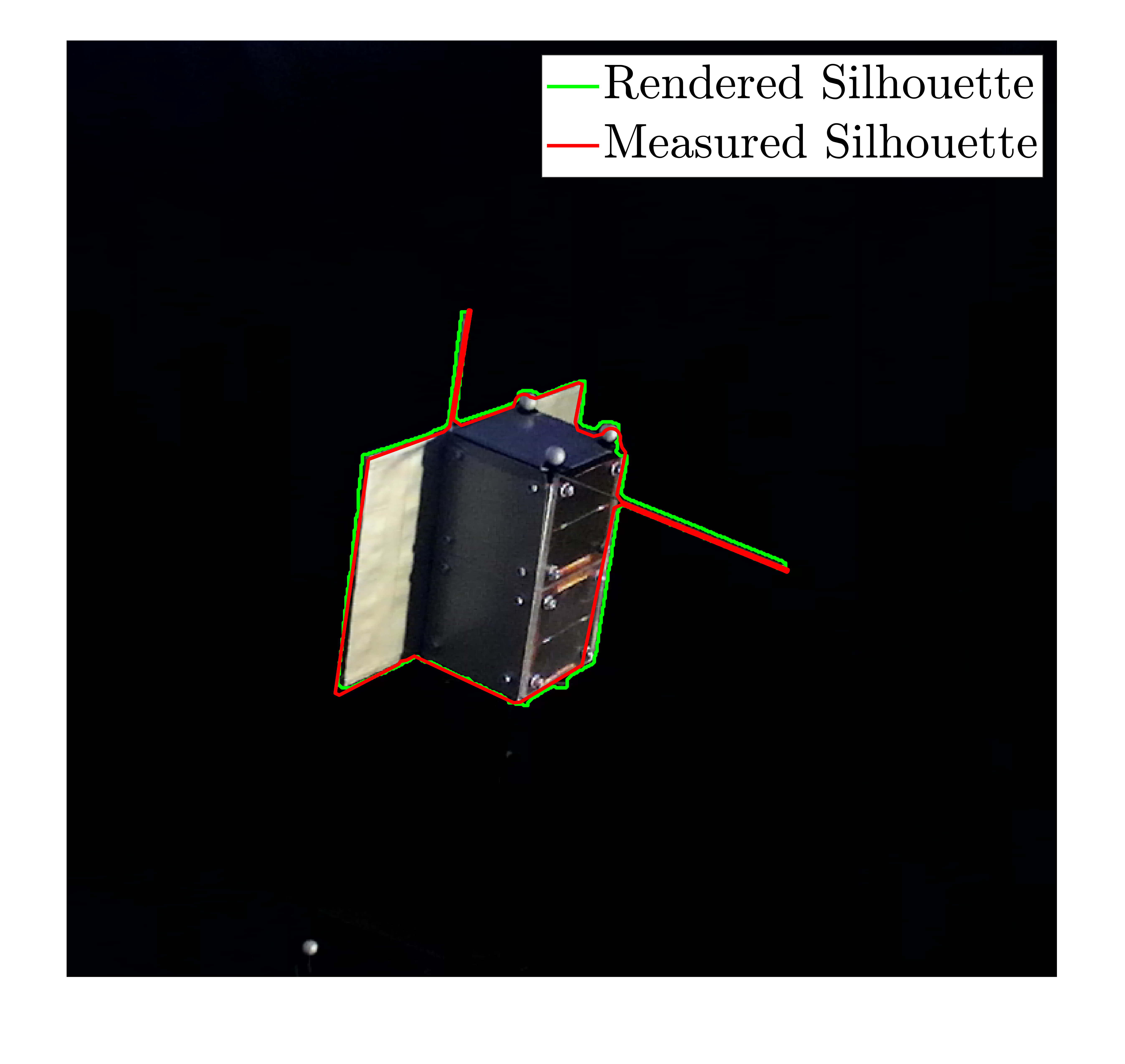}\label{fig:renderedVsMeasuredSilhouette}}
   
    \caption{(a) A set of synthetic views are generated from a series of positions arranged on a sphere centered on the CubeSat CAD model. The contours are extracted by performing a grey-scale thresholding, for each silhouette the image moments are calculated. (b) Alignment of the measured silhouette to the reference synthetic model view. (c) Comparison between the rendered and the measured silhouettes corresponding to the ground truth position.}
    
\end{figure*}

\section{Evaluated Pose Initialization Methods}
\label{sec:methods}

The two close-range methods analysed in this work are described in this section. Both methods are based on the knowledge of the target geometry, it is not necessary the use of fiducial markers and any a priori range measurements or state information. The processing flow are highlighted in \figurename{\ref{fig:methodsSchemes}}.

For both methods, starting from the correspondence between a set of 2D image features $\mathbf{x}_{i,j}$   with the 3D geometric model of the target CubeSat $^{W}\mathbf{X}_j$, it is possible to estimate the i-th camera pose $\mathbf{T}_i^W$ relative to the target CubeSat  by solving a Perspective-n-Point problem. The correspondence is built respectively upon the j-th line ends for the SVD method and the j-th point that composes the silhouette edge for the Silhouette-matching method. $\mathbf{T}_i^W$ is the transformation matrix that embeds the attitude $R_i^W$ and position vector $\mathbf{t}_i^W$ of the camera pose. The camera pose of the i-th image $\mathbf{T}_i^W$ is refined by optimizing the following cost with the Levenberg-Marquardt algorithm:
\begin{equation}
^*T_i^W = \argmin_{\mathbf{T}_i^W}\sum_{j=1}^{n}||\mathbf{x}_{i,j}-\pi(^{W}\mathbf{X}_j,_{W}\mathbf{T}_i^W)||
\label{optimization}
\end{equation}
where $\pi()$ is the pinhole projection function. The optimization problem is initialised solving the perspective-three-point (P3P) algorithm \cite{gao2003complete} embedded in a RANdom SAmple Consensus (RANSAC) scheme. Only inliers matches that survive RANSAC are optimized with the problem defined in (\ref{optimization}).

\subsection{Sharma-Ventura-D'Amico method}

The method is based on target SC edges recognition and matching with its wireframe model, see \figurename{\ref{fig:satelliteWireFrame}}. As first, an initial pre-processing procedure is performed on the raw images to correct them for lens distortion and the noise magnitude. Then, the pre-processed images are given as input to the Weak Gradient Elimination (WGE) block, which is entitled to identify the target Region of Interest (ROI). Gradient magnitude thresholding are used to eliminate the weakest elements in the image. Thereafter, the SVD image processing algorithm employs a feature extraction
architecture based on merging two separate streams of data in order to detect different
elements of the image and provide more robustness to the results. The first stream of features is extrapolated from the filtered gradient image
by means of a Hough transform. A second stream of features is obtained with the application of the Sobel operator to the undistorted unfiltered image. Once the edges in the image are identified, the Hough transform is implemented again
to identify the lines that belong to the target vehicle, see \figurename{\ref{fig:recognizedSatellite}}.
 Finally, the feature detection phase is concluded by merging the two streams of features and combining the results.

To reduce the number of associated features, the perceptual grouping method has been applied. Perceptual grouping is the
process of synthetizing all edges detected in the image into higher-level features by
checking the validity of geometric relations among the segments. The edges are divided into six high-level groups which are in order of decreasing
complexity: (1) Polygonal tetrads, (2) Polygonal triads, (3) Parallel triads, (4) Parallel pairs, (5) Proximity pairs, (6) Antennas. The process of feature selection and synthetization is carried on by examining a few geometric constraints regarding orientation, distance, and length. 
Since most of the high-level features do not provide enough points to obtain a unique solution to the PnP problem, only a combination of at least two of them is required to feed as input to the pose solver.  As a set of correspondences are retrieved between the 3D geometric model points and the 2D ends of the geometric features identified in the image. The target CubeSat pose is finally retrieved solving the camera resectioning problem as defined in Eq. \ref{optimization}.

\subsection{Silhouette matching method}

The method is based on the alignment between the silhouette extracted from the current target view and a candidate synthetic view of the target obtained off-line, see \figurename{\ref{fig:shiluhetteGeneration}}. The synthetic views $\mathbf{s}_i$ are generated starting from the target CAD and placing the camera views on a sphere centered on the 3D target model at spaced points with spherical coordinates $(\alpha_i, \beta_i, \rho_i)$.

The correct correspondence is retrieved by an exhaustive search based on silhouette similarity. The pre-processing block is similar to that seen for the SVD method (distortion correction and ROI identification). Then, edge detection with Sobel operator and morphological operations (black and white dilation, flood-fill operation,  light structures connected to image border suppression, and image erosion) are used to extract the SC silhouette from the ROI.

The correspondence between the current silhouette and the database is retrieved with a greedy search based on silhouette similarity. Silhouette matching is retrieved by computing the    \textit{shape context} between the two silhouette (see \cite{belongie2000shape}).  The target CubeSat pose is finally retrieved solving the camera resectioning problem as defined in Eq. \ref{optimization}.

\begin{figure}[h!]
    \centering
   
        \subfloat[]{\includegraphics[width=0.22\textwidth]{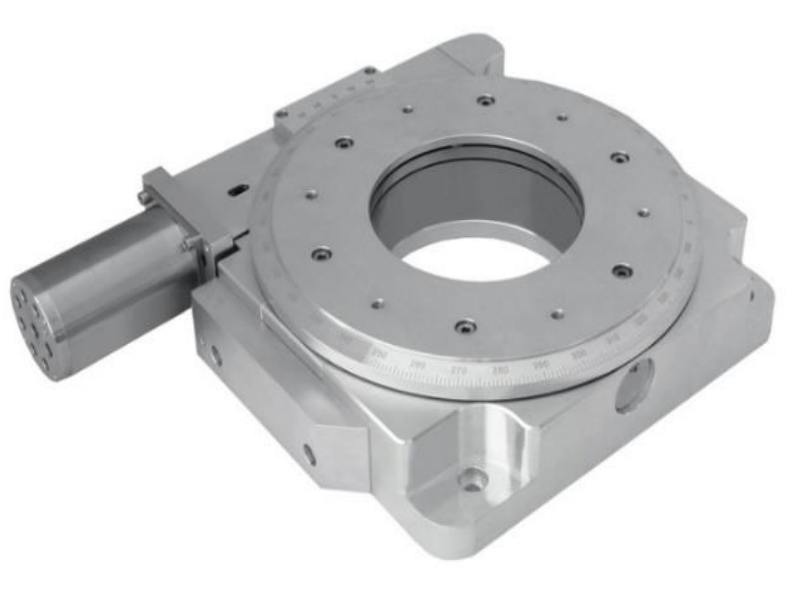}\label{Fig:RotaryStage}}
        \hfill
        \subfloat[]{\includegraphics[width=0.22\textwidth]{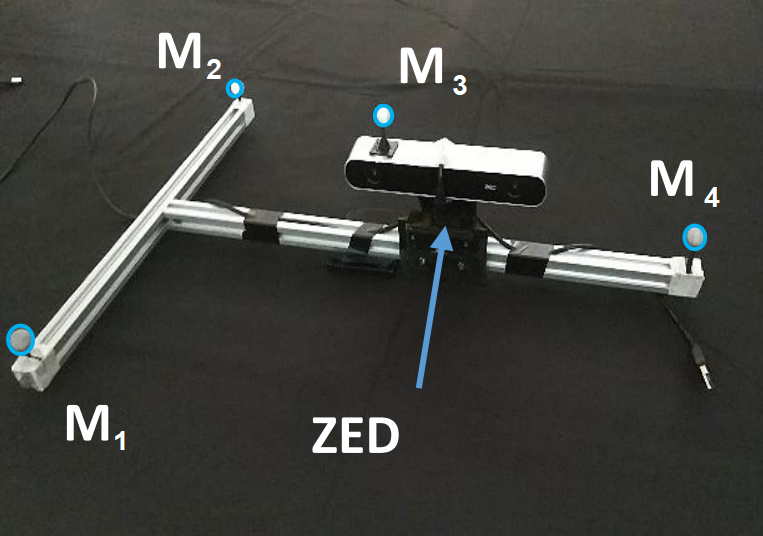}\label{Fig:reflectiveMarkers}}
        \vfill
        \subfloat[]{\includegraphics[width=0.44\textwidth]{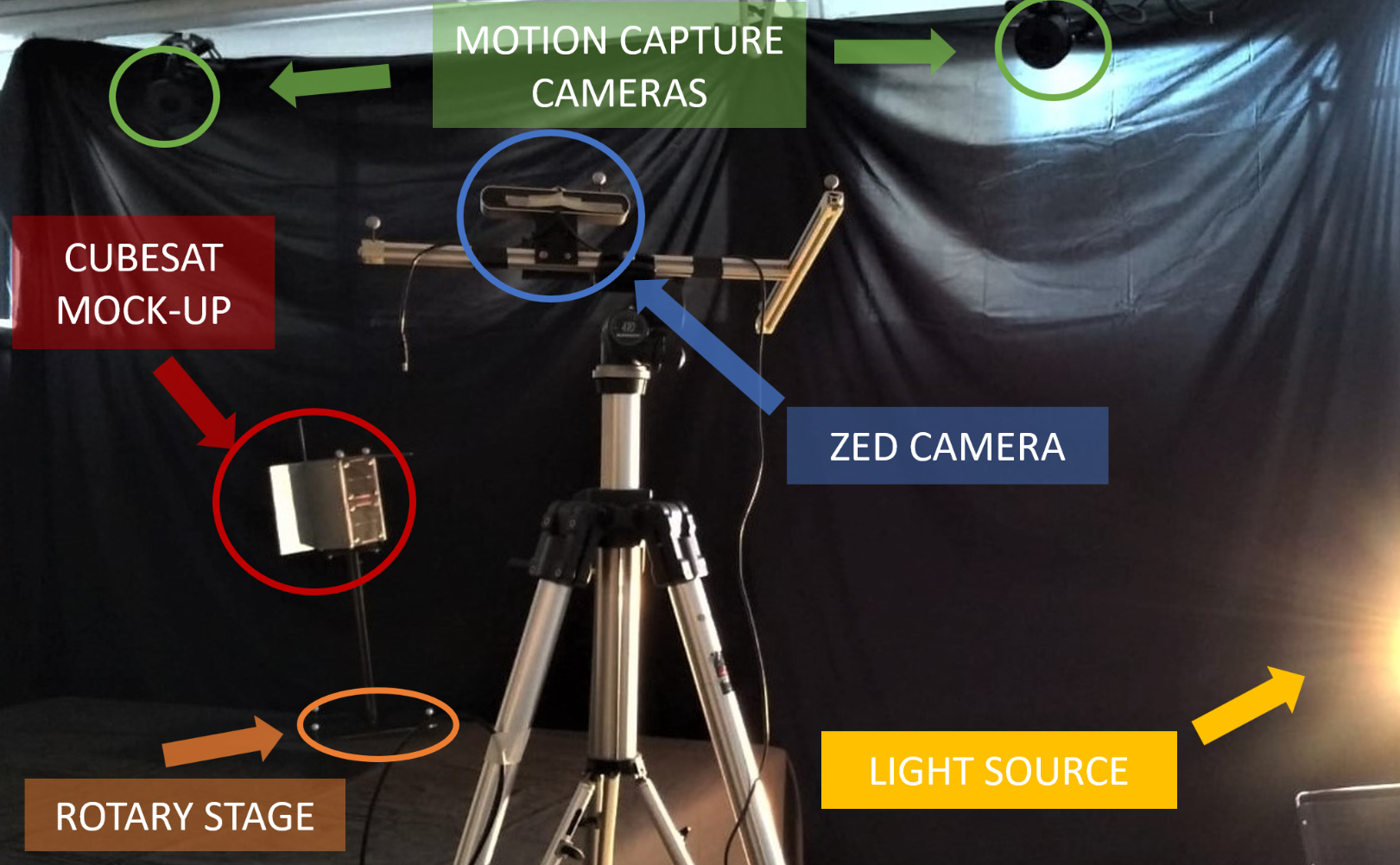}\label{Fig:experimentalSetUp}}
   
    \caption{(a) A rotary stage has been used to rotate the CubeSat by regular steps. (b) ZED camera mounted on its own support frame with four reflective markers used for motion tracking. (c) Experimental setup and lighting conditions adopted in the laboratory. 
    }

\end{figure}

\section{Experimental Setup}

To test the pose initialization capabilities of the two methods, a video sequence was created that simulates the orbiting of a chaser spacecraft around a target CubeSat. The satellite mock-up consisted in a simplified 1:1 model of a 2U-CubeSat, a common format for miniaturized satellites widely used in cost-effective Low Earth Orbit (LEO) missions. The model was placed at a distance equal to 1630 mm from the camera on a high precision motorized rotary stage, see \figurename{\ref{Fig:RotaryStage}}. During the images acquisition, the rotary stage was controlled to impose a full 360$^{\circ}$ rotation on the CubeSat around the vertical axis with a rotation speed of 1.5 ${\circ}/s$; moreover, the acquisition rate was set to acquire a frame every 0.84$^{\circ}$. The axis of rotation was rotated 30 degrees with respect to the longitudinal side of the CubeSat.  The camera setup consisted in a ZED stereo-camera (110$^\circ$ FOV, $1920\times1080$ px) system rigidly mounted on an aluminium frame attached to an adjustable tripod. The sequence consists of 475 images.

Both the camera and the CubesSat were equipped with a set of retro-reflective spherical markers that were tracked during all the acquisitions by a motion capture system with millimetre accuracy. The markers attached to the mock-up of the CubeSat determine its reference system, while, concerning the camera, the retro-reflective markers were employed to track the supporting frame of the camera. Therefore, it was necessary to carry out a calibration procedure to obtain the transformation between the optical reference frame and the camera support frame. The relative pose has been calculated by minimizing the reprojection error on the acquired images of a set of retro-reflective markers placed in the center of the scene and visible both from the motion capture and the camera. The following procedure, developed by \cite{chiodini2019experimental}, has been used: (1) retro-reflective marker detection on acquired images and 2D center position computation by means of Circular Hough Transform; (2) association of the 3D retro-reflective marker to their 2D image using a first rough approximation of the relative pose; (3) camera pose estimation with reference to the global motion capture reference frame through a Perspective-3-Point (P3P) algorithm embedded inside a RANdom SAmple Consensus (RANSAC) scheme, and then refinement with global optimization; (4) reference system relative pose estimation. The application of this method to the motion capture system used in this work showed that it is possible to reconstruct the position of the camera with millimeter accuracy, and alignment better than 0.8 deg (with a confidence level of 95.45\%), as shown in \cite{valmorbida2020calibration}. 

\figurename{\ref{Fig:reflectiveMarkers}} shows the ZED camera mounted on its own support frame with four reflective markers used for motion tracking. \figurename{\ref{Fig:experimentalSetUp}} shows the entire experimental set-up, including the CubeSat equipped with retroreflective markers to measure its motion and the position of the lamp used to simulate the lighting conditions.

\begin{figure}[h!]
    \centering

    \subfloat[Sharma-Ventura-D'Amico Method]{\includegraphics[width=0.45\textwidth]{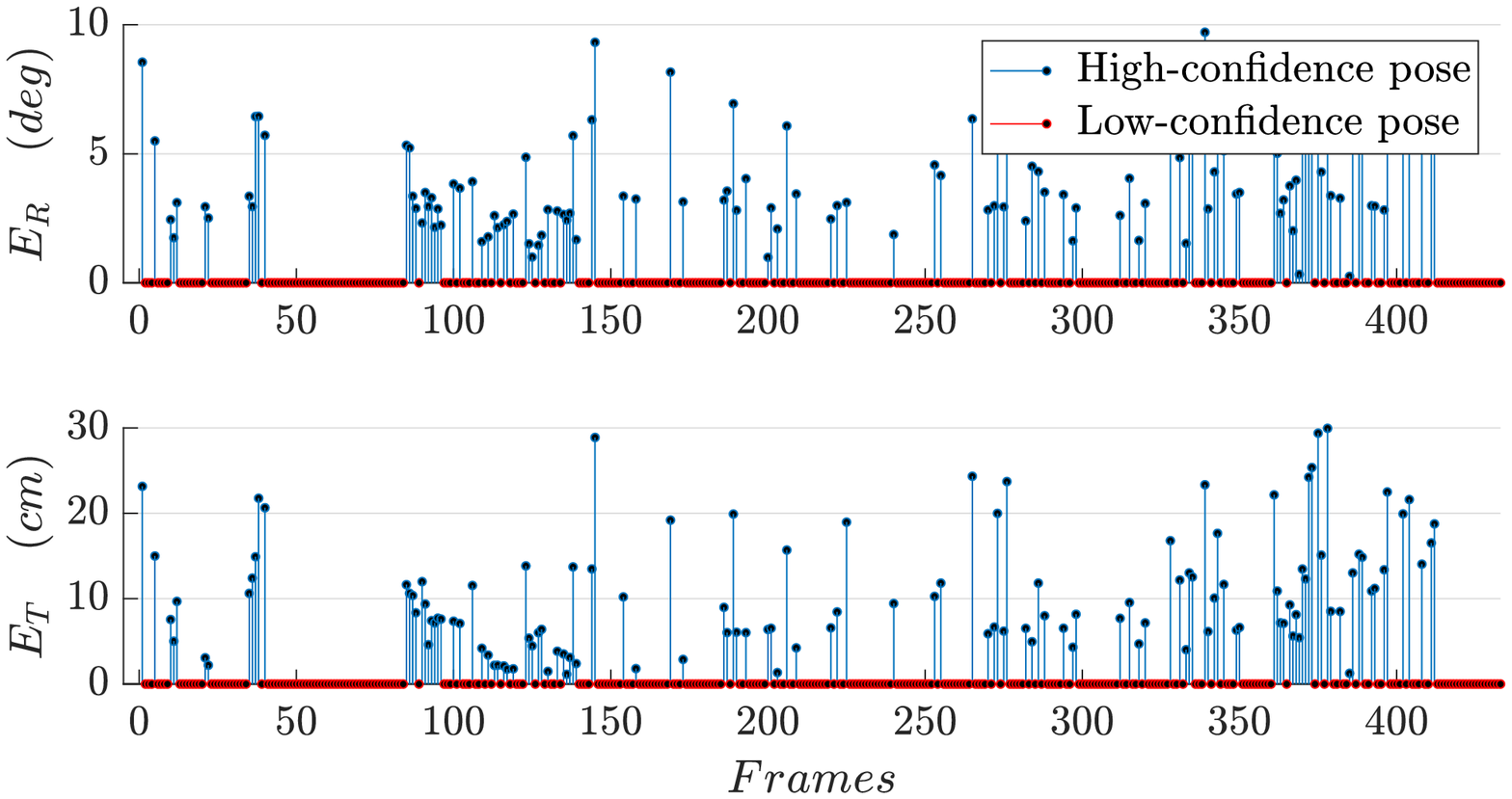}\label{Figresults}}
    \\
    \subfloat[Silhouette matching method]{\includegraphics[width=0.45\textwidth]{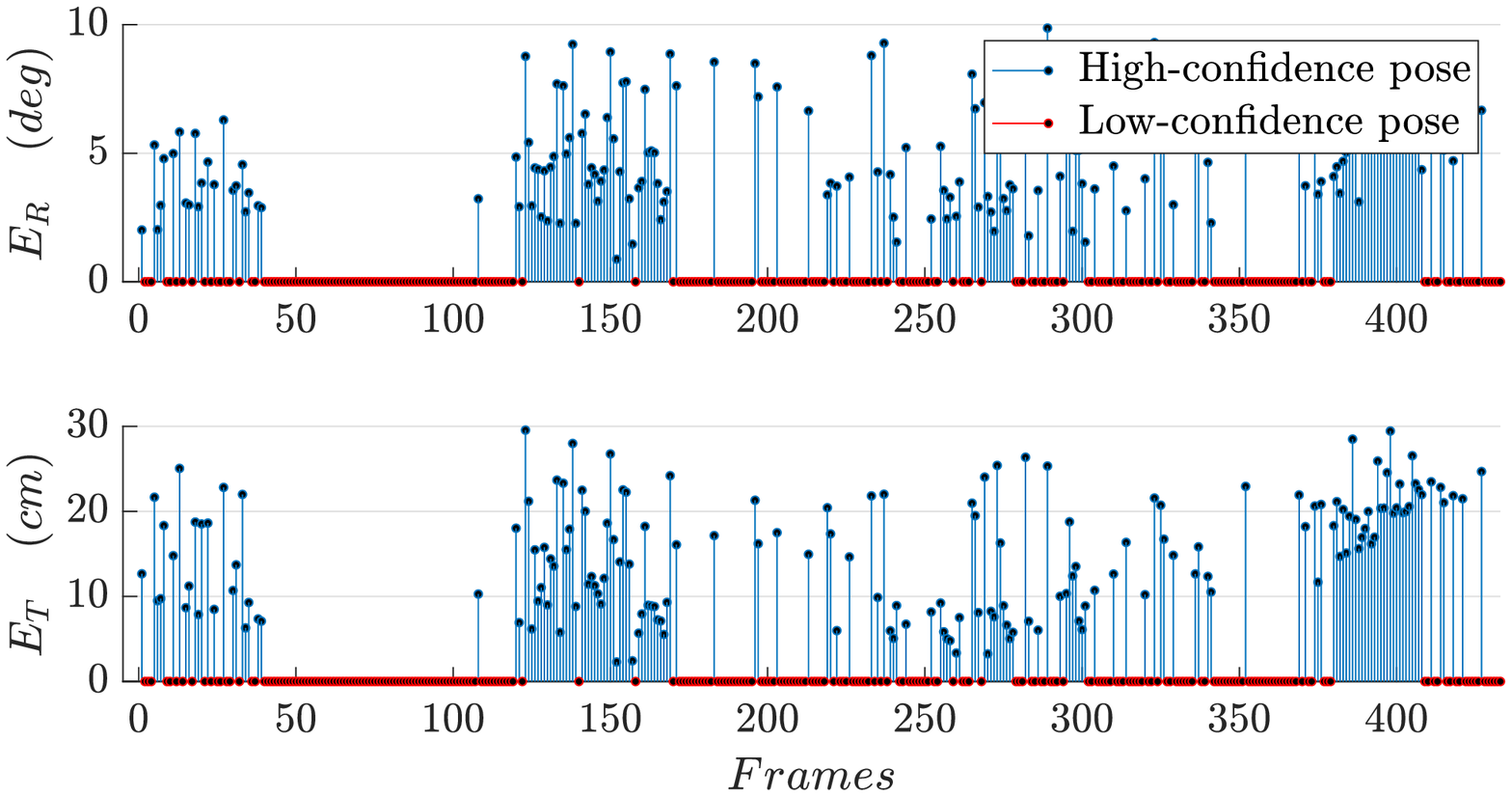}\label{Figresults}}
        \\
    \subfloat[]{\includegraphics[width=0.4\textwidth]{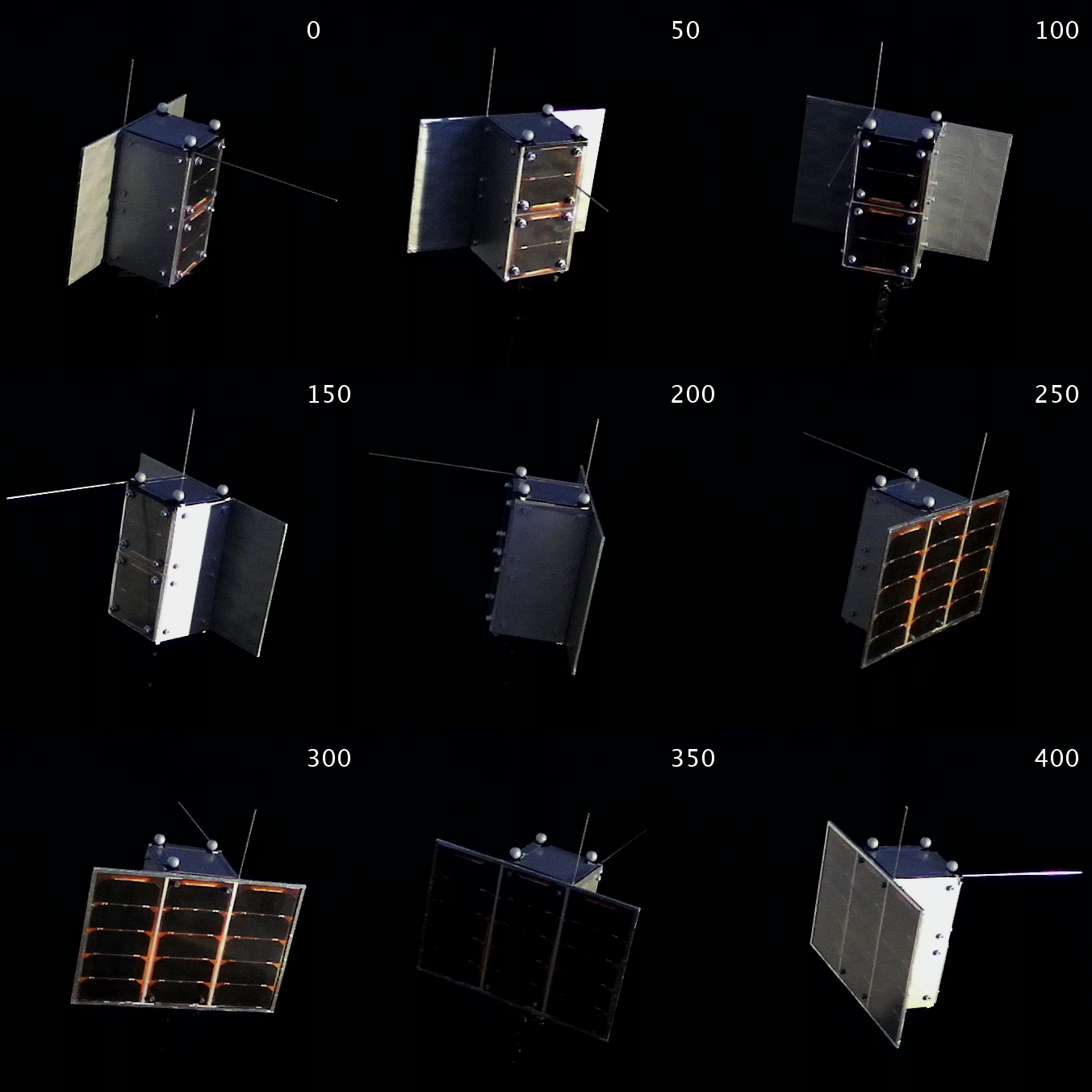}\label{Fig:satelliteImages}}

    \caption{(a-b) Pose reconstruction error for evaluated frames. Only poses reconstructed with $E_T <30$ cm and $E_{R}<10$ deg are considered successfully reconstructed. (b) Images used as input to the algorithm cropped around the target CubeSat with WGE ROI, which is around $450\times450$ pixels.}
    \label{fig:results}
\end{figure}

\begin{figure}[h!]
    \centering

    \subfloat[Sharma-Ventura-D'Amico Method]{\includegraphics[width=0.45\textwidth]{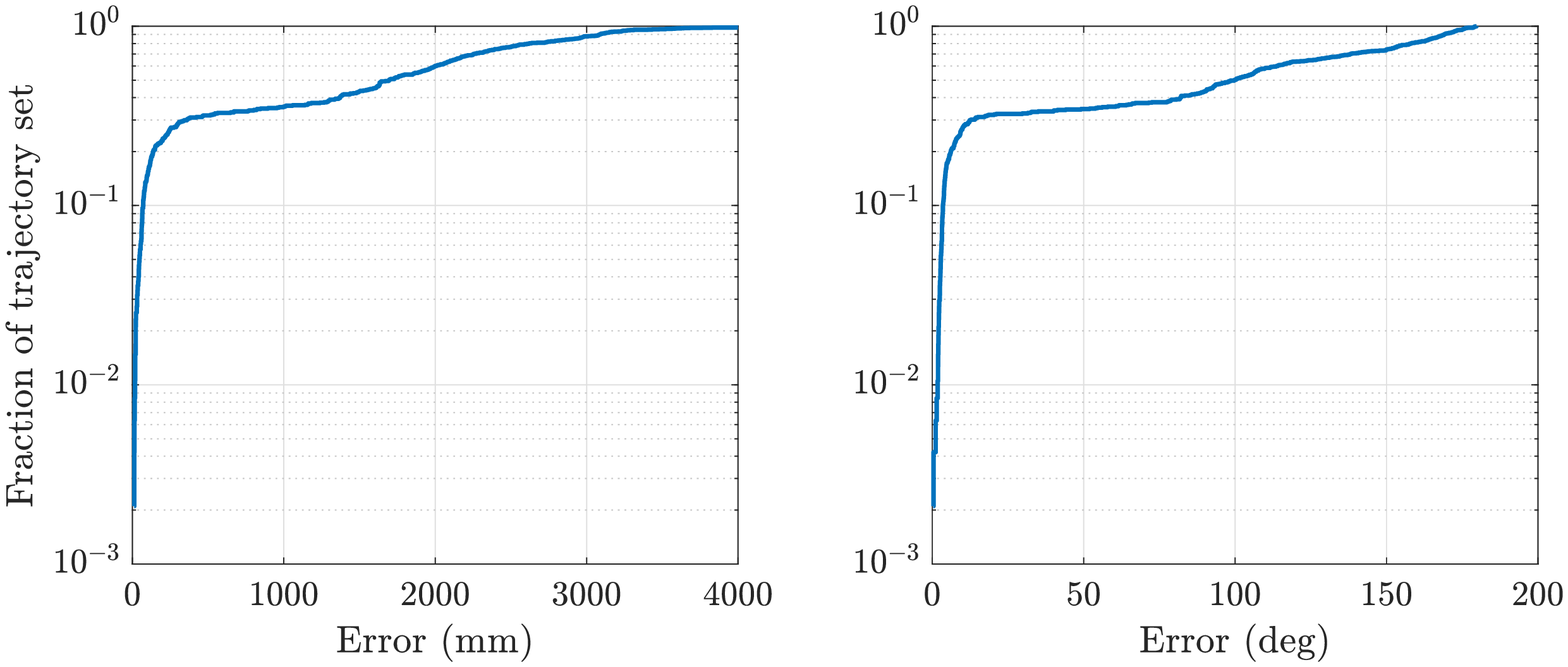}\label{}}
    \\
    \subfloat[Silhouette matching method]{\includegraphics[width=0.45\textwidth]{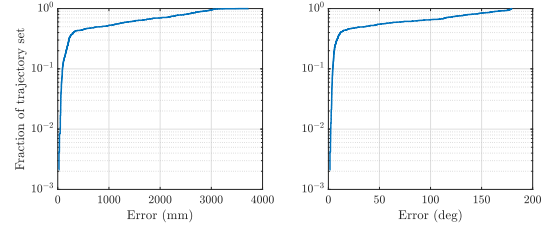}\label{}}

    \caption{ Cumulative distribution function for the position and attitude errors. As can be seen, an initial steep increase in both curves highlights that 27\% - for the Sharma-Ventura-D’Amico Method method (a) - and 35\% - for the Silhouette matching method (b) - of the images are characterized by relatively low pose errors.}
    \label{fig:cfd}
\end{figure}

\section{Results}
\label{sec:results}

To evaluate the accuracy of the two methods analyzed in this paper in estimating the relative pose between two satellites, the Absolute Trajectory Error (ATE) and rotation error metrics have been used. The ATE is defined as the L2 norm of the vector difference between the true and the estimated positions:
\begin{equation}
	e_{t} = ||\mathbf{p}_i - \mathbf{p}_i^*||_2
\end{equation}
where $\mathbf{p}_i^*$ denotes ground truth position at frame $i$, as measured by the motion capture, and $\mathbf{p}_i$ denotes the one estimated by the algorithm. The rotation error is given by:
\begin{equation}
	e_{\theta}=\arccos{(\frac{\Tr(\mathbf{R_i}^{\top}\mathbf{R_i^*})-1}{2})}
\end{equation}
where $\mathbf{R_i^*}$ denotes ground truth rotation matrix at frame $i$, as measured by the motion capture, and $\mathbf{R_i}$ denotes the one estimated by the algorithm.
This metric tells how much the trajectory is deviating from the true path. 

The Silhouette matching method requires the generation of a database of silhouettes before the run of the pose reconstruction algorithm. The grid on which the rendered silhouettes were generated was made at a distance $r = [1500\quad1600\quad1700]$ mm, inclinations $\beta = [-30\quad-20\quad-10\quad0\quad10\quad20\quad30]$ and angular step $\Delta\alpha=10$ deg. A preliminary estimate of the target CubeSat distance from the camera can be obtained from the size of the WGE ROI.

\figurename{\ref{fig:results}} (a-b) show the pose reconstruction error for evaluated frames, in terms of ATE error and rotation error. Only poses reconstructed with $E_T <30$ cm and $E_{R}<10^{\circ}$ are considered successfully reconstructed. Tab. \ref{tab:results} shows the mean error of the successfully reconstructed poses. \figurename{\ref{fig:results}} (c) shows the images used as input to the algorithm cropped around the target CubeSat. It should be noted that the CubeSat occupies a marginal part of the FOV of the camera, only 10\%, despite this the analyzed methods manage to converge in a significant portion of cases towards a solution characterized by a relatively small pose error. \figurename{\ref{fig:cfd}} shows cumulative distribution function for the position and attitude errors. As can be seen, an initial steep increase in both curves highlights that 27\% - for the Sharma-Ventura-D’Amico Method method - and 35\% - for the Silhouette matching method - of the images are characterized by relatively low pose errors ($E_T <30$ cm and $E_{R}<10^{\circ}$). Even if the percentages of success may seem low, we would like to highlight that these methods are used for pose initialization. These methods would be practiced during the inspection phase of the target CubeSat. The results show that by making one orbit around the target there are 27\% - for SVD - and 35\% - for the silhouette matching method - of images useful to initialize the pose for subsequent relative pose tracking.

	\begin{table}[ht]

\centering
\caption{Comparison between SVD and Silhouette matching methods, mean error of the successfully reconstructed poses.}

\begin{tabular}[t]{l|c|c}
\toprule
&SVD*  & Silhouette Matching**\\
\midrule
Mean $E_T$ [cm] & 10 & 14\\
Mean $E_R$ [deg] & 4.7 & 5.7\\
\bottomrule
\end{tabular}

\begin{tablenotes}
      \small
      \item (*) Mean over 126 samples (**) Mean over 167 - Total samples: 475.
    \end{tablenotes}
\label{tab:results}
\end{table}%

\section{Conclusions}
\label{sec:conclusions}

In this work, the experimental comparison between two methods for initializing the pose of a target CubeSat with respect to a chaser camera was carried out. The methods tested are the Sharma-Ventura-D'Amico method and the silhouette matching method. The results show
that by making one orbit around the target there are 27\% -
for SVD - and 35\% - for Silhouette matching method - of
images useful to initialize the target CubeSat pose relative to the chaser camera. The preliminary results show that the average position error is 10 cm for the SVD method and 14 cm for the silhouette matching method in the tested configurations, while the rotation error is, respectively, $4.7^{\circ}$ and $5.7^{\circ}$. Future work will address methods to detect low-confidence cases without ground-truth, as an example by checking the number of matched points and the optimization error.

\bibliographystyle{IEEEtran}
\bibliography{IEEEabrv,bare_conf}

\end{document}